\documentclass[graybox]{svmult}

\usepackage{mathptmx}       
\usepackage{helvet}         
\usepackage{courier}        
\usepackage{type1cm}        
\usepackage{amsmath}
\usepackage{amssymb}
\usepackage{makeidx}         
\usepackage{graphicx}        
\usepackage{pdfsync}                             
\usepackage{multicol}        
\usepackage[bottom]{footmisc}

\newcommand{\bfmu}{{\boldsymbol{\mu}}}
\newcommand{\bfh}{{\textbf{h}}}

\newcommand{\bfx}{{\textbf{x}}}

\newcommand{\bfu}{{\textbf{u}}}
\newcommand{\bfw}{{\textbf{w}}}

\newcommand{\bfg}{{\textbf{g}}}

\makeindex

\begin{document}

\title*{Learning manifold to regularize nonnegative matrix factorization}

\author{Jim Jing-Yan Wang and Xin Gao}

\institute{Jim Jing-Yan Wang \and Xin Gao \at
Computer, Electrical and Mathematical Sciences and Engineering Division, King Abdullah University of Science and Technology (KAUST), Thuwal 23955-6900, Saudi Arabia \\
\email{jimjywang@gmail.com}, \email{xin.gao@kaust.edu.sa}}

\maketitle

\abstract{In this chapter we discuss how to learn an optimal manifold presentation to regularize nonegative matrix factorization (NMF) for data representation problems. NMF, which tries to represent a nonnegative data matrix as a product of two low rank nonnegative matrices, has been a popular method for data representation due to its ability to explore the latent part-based structure of data. Recent study shows that lots of data distributions have manifold structures, and we should respect the manifold structure when the data are represented. Recently, manifold regularized NMF used a nearest neighbor graph to regulate the learning of factorization parameter matrices and has shown its advantage over traditional NMF methods for data representation problems. However, how to construct an optimal graph to present the manifold properly remains a difficult problem due to the graph model selection, noisy features, and nonlinear distributed data. In this chapter, we introduce three effective methods to solve these problems of graph construction for manifold regularized NMF. Multiple graph learning is proposed to solve the problem of graph model selection, adaptive graph learning via feature selection is proposed to solve the problem of constructing a graph from noisy features, while multi-kernel learning-based graph construction is used to solve the problem of learning a graph from nonlinearly distributed data.}

\section{Introduction}
\label{sec:intro}

Nonnegative matrix factorization (NMF) \cite{Kopriva201447,li2014graph,qingquan2010context,sun2012unsupervised,sun2014mobile,sun2014non} has been a popular data representation method in field of machine learning. Given a matrix of nonnegative data, where each column is a data sample, NMF tries to represent it as a product of two low rank nonnegative matrices, i.e., a basis matrix and a coefficient matrix. The coefficient matrix could be used as new representations of the data samples. Due to its ability to explore the latent part-based structure of the data, it has been widely used to many real-world applications, such as computer vision \cite{Meganem20141822},
optical sensing \cite{Cui10} and bioinformatics \cite{wang2013beyond,wang2013non}.

To formulate the problem of NMF in a matrix form, we assume that we have a data set of $n$ data samples, and their feature vectors are $\{\bfx_1,\cdots,\bfx_n\}$, where $\bfx_i\in R^d_+$ is a $d$-dimensional nonnegative feature vector of the $i$-th sample. The data set is organized as a nonnegative matrix $X=[\bfx_1,\cdots,\bfx_n] \in R_+^{d\times n}$. The NMF problem is to learn two nonnegative matrices $H \in R_+^{d\times m}$ and $W\in R_+^{m\times n}$ so that the original matrix $X$ can be reconstructed as the product of $H$ and $W$,

\begin{equation}
\begin{aligned}
X \approx H W.
\end{aligned}
\end{equation}
In $H$ each column can be regarded as a basis vector. $m$ is the number of basis vectors and usually $m\ll n$. In this way, each $\bfx_i$ can be reconstructed as the linear combination of the basis vectors in $H$ and the $i$-th column in $W$ is the combination coefficient vector,

\begin{equation}
\begin{aligned}
\bfx_i \approx H \bfw_i,
\end{aligned}
\end{equation}
where $\bfw_i \in R_+^m$ is the $i$-th column of $W$, and it can be viewed as a low-dimensional nonnegative representation of $\bfx_i$. To learn both the matrices $H$ and $W$, the reconstruction error is usually measured by a squared $\ell_2$ norm distance between $X$ and $HW$,

\begin{equation}
\label{equ:NMF}
\begin{aligned}
\min_{H,W}
~ &\|X - H W\|_2^2\\
s.t.
~& H \geq 0, W \geq 0.
\end{aligned}
\end{equation}
This problem can be solved by the Lagrange multiplier method \cite{lee2001algorithms}. Some other functions can also be used as reconstruction error measures, such as Kullback - Leibler divergence \cite{Hughes2014IV}, earth mover's distance \cite{Wang2014}, etc.

Recent studies on manifold learning \cite{Ding20141987} show that lots of data distributions have low-dimensional manifold structures. The manifold structure has been explored by component analysis methods \cite{Roweis20002323,He20145286}, ranking score learning methods \cite{zhou2004ranking,wang2012multipleranking}, etc.
It usually constructs a nearest neighbor graph from the original feature space to present the manifold structure of the data set, and then use the affinity matrix of the graph to regularize the learned outputs from the data. Given a data sample $\bfx_i$, we denote its nearest neighbor set as $\mathcal{N}_i$. A graph is constructed as $G=(\mathcal{X}, \mathcal{E}, A)$, where $\mathcal{X} = \{\bfx_1,\cdots,\bfx_n\}$ is the node set of the graph and each node is a data point, $\mathcal{E}$ is the edge set, and $A \in R^{n\times n}$ is the affinity matrix of the graph. $\mathcal{E}$ is defined as

\begin{equation}
\begin{aligned}
\mathcal{E} = \{(\bfx_i,\bfx_j)| \bfx_i,\bfx_j \in \mathcal{X}, ~and~\bfx_j \in \mathcal{N}_i\}
\end{aligned}
\end{equation}
and $A$ is defined as

\begin{equation}
\label{equ:affinity}
\begin{aligned}
A_{ij}=
\left\{\begin{matrix}
g(\bfx_i,\bfx_j), &if~(\bfx_i,\bfx_j) \in \mathcal{E}, \\
0, & otherwise.
\end{matrix}\right.
\end{aligned}
\end{equation}
where $g(\bfx_i,\bfx_j) = exp\left (\frac{\|\bfx_i - \bfx_j\|_2^2}{2 \sigma_{ij}^2}\right)$ is a Gaussian kernel function and $\sigma_{ij}$ is the band-width parameter.
Cai et al. \cite{cai2011graph} argued that when the data are represented using NMF, the manifold structure  should also be respected and they used the graph affinity matrix to regularize the learning of the coefficient matrix $W$. It is imposed that if two data samples, $\bfx_i$ and $\bfx_j$, are connected in the graph and their affinity is large, their coefficient vectors should also be close to each other. They used the squared $\ell_2$ norm distance to measure how close they are from each other, and proposed the following minimization problem,

\begin{equation}
\begin{aligned}
\min_{W} & \frac{1}{2} \sum_{i,j=1}^n A_{ij} \|\bfw_i - \bfw_j \|_2^2\\
s.t& W \geq 0.
\end{aligned}
\end{equation}
In this way, the local manifold information is mapped from the original space to the coefficient vector space by minimizing the pairwise coefficient vector  distance weighted by the affinity calculated from the original feature space. This problem is combined with the original NMF problem in (\ref{equ:NMF}) and the following graph regularized NMF problem is obtained,

\begin{equation}
\label{equ:GrNMF}
\begin{aligned}
\min_{H,W}
~ &\|X - H W\|_2^2 + \alpha \frac{1}{2} \sum_{i,j=1}^n A_{ij} \|\bfw_i - \bfw_j \|_2^2\\
s.t.
~& H \geq 0, W \geq 0,
\end{aligned}
\end{equation}
where $\alpha$ is a tradeoff parameter.

However, on the construction of the graph, the following problems remain difficult, which prevent the graph regularization from being widely used to NMF problems:

\begin{enumerate}
\item To find the nearest neighbors of a data sample, we need to determine which distance function is optimal. Moreover, the size of the neighborhood should also be decided.
\item To calculate the affinity measure between a pair of data samples, we should select an optimal bankwidth parameter for the Gaussian kernel in (\ref{equ:affinity}). Moreover, besides Gaussian kernel function, we can use some other affinity measures to calculate the affinity matrix. Thus it is also necessary to select an optimal affinity function.
\item If some features extracted from the data samples are noisy or irrelevant to the problem on hand, the graph constructed from the original feature space is not suitable to regularize the NMF learning. In this case, usage of original features to construct the graph will harm the learning performance.
\item If the data distribution of the data set is not linear, and we still use the linear functions to construct the graph, the graph will not be optimal for learning of NMF. In this case, we should use kernel tricks \cite{Mariethoz20072315} to map the data to a nonlinear space first and then construct graph.
\end{enumerate}

To solve these problems, in the following sections we will introduce some varieties of graph regularized NMF. In Section \ref{sec:MultiGraph}, we introduce the multiple graph regularize NMF to solve the first two problems by learning an optimal graph from some candidate graphs with different distance functions, graph models and parameters. In Section \ref{sec:FS}, to solve the third problem, we introduce a method to integrate feature selection to NMF and also use it to refine the graph for regularization of NMF.  In Section \ref{sec:MultiK}, we introduce a method to combine multi-kernel learning and NMF, and use it to refine the graph in the kernel space to solve the fourth problem.

\section{Multiple graph regularized NMF}
\label{sec:MultiGraph}

To construct a graph for the regularization of NMF, we need to select a distance function and a neighborhood size to determine the nearest neighbors of a given data sample. Moreover, we also need to select an affinity function and its corresponding parameters to calculate the affinities between each pair of connected data samples. We can use a linear search in the hyper space of distance function, affinity function, and parameters to find the optimal combination of choices to construct the graph. However, this strategy is time-consuming and easy to be over-fitting to the training data set. To overcome this problem, in our previous work \cite{wang2013multiple}, we proposed the multiple graph learning method to construct an optimal graph for regularization of NMF.

We assume we have $l$ different candidate graphs and they are constructed with combinations of different distance functions, affinity functions, and parameters. Their affinity matrices are denoted as $\{A^1, \cdots, A^l\}$, where $A^k \in R_+^{n \times n}$ is the affinity matrix of the $k$-th graph. The problem is to learn an optimal graph affinity matrix $A^*$ from them. To solve this problem, we assume that the optimal graph affinity matrix can be obtained by the linear combination of the candidate graph affinity matrices,

\begin{equation}
\label{equ:A}
\begin{aligned}
&A^* = \sum_{k=1}^l \mu_k A^k,\\
&s.t. \sum_{k=1}^l \mu_k = 1, \mu_k \geq 0,
\end{aligned}
\end{equation}
where $\mu_k$ is the linear combination weight of the $k$-th graph affinity matrix. The constrains $\sum_{k=1}^l \mu_k = 1$ and $\mu_k \geq 0$ are imposed to prevent the negative weights. In this way, we transfer the problem of selecting an optimal graph from a pool of candidate graphs to a problem of learning the combination weights. To learn the combination weights, we integrate (\ref{equ:A}) to (\ref{equ:GrNMF}), and obtain the following minimization problem,

\begin{equation}
\label{equ:MultiGrNMF}
\begin{aligned}
\min_{H,W, \mu_k|_{k=1,\cdots,l}}
~ &\|X - H W\|_2^2 + \alpha \frac{1}{2} \sum_{i,j=1}^n \sum_{k=1}^l \mu_k A^k_{ij} \|\bfw_i - \bfw_j \|_2^2 + \beta \sum_{k=1}^l \mu_k^2\\
s.t.
~& H \geq 0, W \geq 0,\\
& \sum_{k=1}^l \mu_k = 1, \mu_k \geq 0.
\end{aligned}
\end{equation}
Please note that the last term $\sum_{k=1}^l \mu_k^2$ in the objective function is used to prevent  $A^*$ from over-fitting to one single candidate graph, and $\beta$ is its tradeoff parameter. The optimization of this problem is solved using an alternate optimization strategy in an iterative algorithm. In each iteration, we first fix $\mu_k, k=1,\cdots, l$, and update $H$ and $W$, and then fix $H$ and $W$ to update $\mu_k, k=1,\cdots, l$. The detailed optimization procedures are given as follows:

\begin{itemize}
\item \textbf{Updating $H$ and $W$ while fixing $\mu_k, k=1,\cdots, l$}: When $\mu_k, k=1,\cdots, l$ are fixed, the problem in (\ref{equ:MultiGrNMF}) is reduced to the problem in (\ref{equ:GrNMF}), which can be solved using the Lagrange multiplier method introduced in \cite{cai2011graph}.

\item \textbf{Updating $\mu_k, k=1,\cdots, l$ while fixing $H$ and $W$ }: When $H$ and $W$ are fixed, and only $\mu_k, k=1,\cdots, l$ are considered, the problem in  (\ref{equ:MultiGrNMF}) turns to

\begin{equation}
\label{equ:mu}
\begin{aligned}
\min_{\mu_k|_{k=1,\cdots,l}}
~ &\alpha \frac{1}{2} \sum_{i,j=1}^n \sum_{k=1}^l \mu_k A^k_{ij} \|\bfw_i - \bfw_j \|_2^2 + \beta \sum_{k=1}^l \mu_k^2\\
s.t.
~
& \sum_{k=1}^l \mu_k = 1, \mu_k \geq 0.
\end{aligned}
\end{equation}
This problem is a linearly constrained quadratic programming (QP) problem \cite{Shi2014871,Fomeni2014173,Miao201480} and can be solved easily using the active set algorithm \cite{Leng2013638,Jian2014158,Cheng2014763}.

\end{itemize}

The advantages of this method are two folds:

\begin{enumerate}
\item In our method, no class label or any other supervision information is needed to learn the optimal graph. Traditional methods, such as linear search or cross validation need the supervision information to select the optimal graph. However, in many real-world applications, no supervision information is provided. In our method, we use the NMF objective as a criterion to learn an optimal graph and thus avoid the input of supervision information.
\item The learned graph affinity does not fit to any one single graph, because we used a $\ell_2$ norm term to regularize the learning of graph weights. In this way, the proposed method can explore the graph space to utilize the complementary information provided by different graphs.
\end{enumerate}

\section{Feature selection for adaptive graph regularized NMF}
\label{sec:FS}

Traditional graph construction methods construct a graph from the original feature space of the data samples. However, in many cases, there are many noisy and/or irrelevant features in the feature set and these features will affect the graph \cite{Guo2014408}. Actually, using noisy features and/or irrelevant features of data samples to construct the nearest neighbors and to calculate the affinity matrix will make the obtained graph unsuitable for the problem on hand, or even make the performance poorer than the method without graph regularization. To solve this problem, we proposed to integrate feature selection to NMF and use the selected features to construct the graph in our previous work \cite{wang2012adaptive}. To this end, we proposed to weight each feature with a feature weight and represent a data sample $\bfx_i$ as follows,

\begin{equation}
\label{equ:FS}
\begin{aligned}
&\bfx_i \rightarrow [u_1x_{i1}, \cdots, u_d x_{id}]^\top = diag(\bfu) \bfx_i,\\
& s.t. \sum_{c=1}^d u_c = 1, u_c\geq 0, c=1,\cdots, d,
\end{aligned}
\end{equation}
where $x_{ic}$ is the $c$-th feature of the $i$-th sample, and $w_c$ is the weight of the $c$-th feature. The constrains $\sum_{c=1}^d u_c = 1$  and $u_c\geq 0$ are imposed to prevent negative feature weights. We hope that a feature relevant to the problem on hand can obtain a large feature weight while a noisy feature can be assigned with a small feature weight. We use the same feature weight vector to weight both the original data matrix $X$ and the basis matrix $H$,

\begin{equation}
\label{equ:XFS}
\begin{aligned}
X\rightarrow diag(\bfu) X, ~and~ H \rightarrow diag(\bfu) H.
\end{aligned}
\end{equation}
Moreover, we also calculate the affinity matrix using the weighted feature vectors as

\begin{equation}
\label{equ:AFS}
\begin{aligned}
A_{ij}^\bfu=
\left\{\begin{matrix}
g(diag(\bfu)\bfx_i,diag(\bfu)\bfx_j), &if~(\bfx_i,\bfx_j) \in \mathcal{E}, \\
0, & otherwise.
\end{matrix}\right.
\end{aligned}
\end{equation}
Substituting both (\ref{equ:XFS}) and (\ref{equ:AFS}) to (\ref{equ:GrNMF}), we have

\begin{equation}
\label{equ:NMF_FS}
\begin{aligned}
\min_{H,W,\bfu}
~ &\left \|diag(\bfu)\left (X - H W\right ) \right\|_2^2 + \alpha \frac{1}{2} \sum_{i,j=1}^n A_{ij}^\bfu \|\bfw_i - \bfw_j \|_2^2\\
s.t.
~& H \geq 0, W \geq 0,\\
& \sum_{c=1}^d u_c = 1, u_c\geq 0, c=1,\cdots, d.
\end{aligned}
\end{equation}
In this problem, the feature weight vector $\bfu$ is also a parameter to be optimized together with $H$ and $W$, and the affinity matrix is based on $\bfu$. In this way, the feature weight vector is learned using the criterion of NMF and it is further used to refine the graph to regularize the NMF parameters. To optimize the problem in (\ref{equ:NMF_FS}), we employ the alternate optimization strategy in an iterative algorithm. In each iteration, we first update $A^\bfu$ according to the feature weight vector $\bfu$ in previous iteration, then $H$ and $W$ are optimized using $A^\bfu$, and finally we optimize $\bfu$ by fixing $A^\bfu$, and  $H$ and $W$. The detailed optimization procedures are as follows,

\begin{itemize}
\item \textbf{Updating $H$ and $W$ while fixing $\bfu$ and $A^\bfu$}: When $\bfu$ and $A^\bfu$ are fixed, the problem in (\ref{equ:NMF_FS}) is reduced to

\begin{equation}
\begin{aligned}
\min_{H,W}
~ &\left \|diag(\bfu)\left (X - H W\right ) \right\|_2^2 + \alpha \frac{1}{2} \sum_{i,j=1}^n A_{ij}^\bfu \|\bfw_i - \bfw_j \|_2^2\\
s.t.
~& H \geq 0, W \geq 0.
\end{aligned}
\end{equation}
This problem can also be solved using the Lagrange multiplier method.

\item \textbf{Updating $\bfu$ while fixing $H$, $W$ and $A^\bfu$}: When we only consider $\bfu$ as a variable and fix $H$, $W$ and $A^\bfu$, the problem is reduced to

\begin{equation}
\begin{aligned}
\min_{\bfu}
~ &\left \|diag(\bfu)\left (X - H W\right ) \right\|_2^2\\
s.t.
~&\sum_{c=1}^d u_c = 1, u_c\geq 0, c=1,\cdots, d.
\end{aligned}
\end{equation}
This problem can also be solved as a quadratic programming (QP) problem.

\item \textbf{Updating $A^\bfu$ according to $\bfu$}: After $\bfu$ is updated, we use it to calculate a new affinity matrix $A^\bfu$ as in (\ref{equ:AFS}). Note that the neighborhood of each data sample is also updated in the updated feature space weighted by $\bfu$.

\end{itemize}

\section{Multi-kernel learning for adaptive graph regularized NMF}
\label{sec:MultiK}

In this section, we solve the problem of learning a graph from nonlinearly distributed data. When the data samples are distributed nonlinearly, and we use linear original feature space to find the nearest neighbors and calculate the affinities, the obtained graph will not be suitable for further regularization of NMF. In this case, we can use a nonlinear function to map a data sample $\bfx_i$ from a nonlinear space to a high-dimensional linear space \cite{Kwak20132113}, $\bfx_i \rightarrow \Phi(\bfx_i) \in R^{d'}$, where $d' \geq d$ is the dimension of the high-dimensional linear space, then construct the graph and perform NMF \cite{wang2012multiple}. In this case, we assume that the basis vectors can be obtained by the linear combination of the nonlinear mapping of the training samples,

\begin{equation}
\label{equ:HmultiK}
\begin{aligned}
&\bfh_k = \sum_{i=1}^n g_{ki}\Phi(\bfx_i) = \Phi(X)\bfg_k,\\
& H = \Phi(X)G,
\end{aligned}
\end{equation}
where $\bfh_k$ is the $k$-th basis vector, $\Phi(X) = [\Phi(\bfx_1),\cdots,\Phi(\bfx_n)] \in R^{d'\times n}$, $\bfg_k = [g_{k1},\cdots, g_{kn}]^\top \in R_+^n$ is the nonnegative linear combination weight vector of the $k$-th basis vector, and $G=[\bfg_1,\cdots,\bfg_l] \in R_+^{n\times m}$ is the nonnegative linear combination weight matrix. In this way, the learning of $H$ is transferred to the learning of the parameter matrix $G$. Substituting $\Phi(X)$ and $H$ in (\ref{equ:HmultiK}) to (\ref{equ:NMF}), we obtain the NMF problem in the nonlinear mapping space,

\begin{equation}
\label{equ:NMFmultiK}
\begin{aligned}
\min_{H,G}
~ &\|\Phi(X) - \Phi(X)G W\|_2^2\\
&= Tr\left ( \Phi(X)^\top \Phi(X) \right ) -  2  Tr\left ( \Phi(X)^\top \Phi(X) G W \right )
+ Tr\left ( \Phi(X)^\top \Phi(X) G W W^\top G^\top \right )  \\
s.t.
~& H \geq 0, G \geq 0.
\end{aligned}
\end{equation}
Actually, the nonlinear mapping function $\Phi(\bfx)$ is not defined explicitly, but implicitly via a kernel function which is defined as the inner-product between $\Phi(\bfx_i)$ and $\Phi(\bfx_j)$, $K(\bfx_i,\bfx_j) = \Phi(\bfx_i)^\top \Phi(\bfx_j)$. Moreover, given the training data matrix $X$ and its mapping matrix $\Phi(X)$, we can define the kernel matrix as

\begin{equation}
\label{equ:KernelMat}
\begin{aligned}
&K = \Phi(X)^\top \Phi(X) = [K_{ij}]_{i,j=1,\cdots, n},\\
& K_{ij} = \Phi(\bfx_i)^\top \Phi(\bfx_j) = K(\bfx_i,\bfx_j).
\end{aligned}
\end{equation}
Similar to multiple graph learning, if we have a pool of candidate kernel functions with parameters, we may also learn an optimal kernel function by the linear combination of them,

\begin{equation}
\label{equ:MultiK}
\begin{aligned}
&K=\sum_{k=1}^l \mu_k K^k,\\
&s.t. ~ \sum_{k=1}^l \mu_k= 1,  \mu_k \geq 0, k=1,\cdots,l,
\end{aligned}
\end{equation}
where $K^k$ is the $k$-th kernel and $\mu_k$ is its linear combination weight. Substituting (\ref{equ:MultiK}) to (\ref{equ:NMFmultiK}), we have

\begin{equation}
\begin{aligned}
\min_{H,G,\{\mu_k\}_{k=1,\cdots,l}}
~ & \sum_{k=1}^l \mu_k \left [ Tr\left ( K \right ) -  2  Tr\left (K G W \right )
+ Tr\left ( K G W W^\top G^\top \right ) \right ] \\
s.t.
~& H \geq 0, G \geq 0,\\
& \sum_{k=1}^l \mu_k= 1,  \mu_k \geq 0, k=1,\cdots,l.
\end{aligned}
\end{equation}
Moreover, we also use the combined kernel to measure the affinity between the neighboring data samples,

\begin{equation}
\label{equ:affinity_multikernel}
\begin{aligned}
A_{ij}^\bfmu=
\left\{\begin{matrix}
\sum_{k=1}^l \mu_k K^k_{ij} , &if~(\bfx_i,\bfx_j) \in \mathcal{E}, \\
0, & otherwise,
\end{matrix}\right.
\end{aligned}
\end{equation}
and use it to regularize the learning of the coefficient vectors. The overall optimization problem is obtained as follows,

\begin{equation}
\label{equ:MultiKGNMF}
\begin{aligned}
\min_{H,G,\{\mu_k\}_{k=1,\cdots,l}}
~ & \sum_{k=1}^l \mu_k \left [ Tr\left ( K \right ) -  2  Tr\left (K G W \right )
+ Tr\left ( K G W W^\top G^\top \right ) \right ] \\
&+ \alpha \frac{1}{2} \sum_{i,j=1}^n \left ( \sum_{k=1}^l \mu_k K^k_{ij} \right ) \left \| \bfw_i - \bfw_j \right \|_2^2 + \beta \sum_{k=1}^l \mu_k^2\\
s.t.
~& H \geq 0, G \geq 0,\\
& \sum_{k=1}^l \mu_k= 1,  \mu_k \geq 0, k=1,\cdots,l,
\end{aligned}
\end{equation}
where $\sum_{k=1}^l \mu_k^2$ is added to the objective function to prevent it from over-fitting to one single kernel, and $\alpha$ and $\beta$ are tradeoff parameters. The solution of this problem can also be obtained by using alternate optimization strategy in an iterative algorithm. In each iteration, $H$, $G$ or $\{\mu_k\}_{k=1,\cdots,l}$ are updated in turn while the other ones are fixed.

\begin{itemize}
\item \textbf{Updating $H$ and $G$ while fixing $\{\mu_k\}_{k=1,\cdots,l}$}: When $\{\mu_k\}_{k=1,\cdots,l}$ are fixed, the problem in (\ref{equ:MultiKGNMF}) is reduced to

\begin{equation}
\begin{aligned}
\min_{H,G}
~ & \sum_{k=1}^l \mu_k \left [ Tr\left ( K \right ) -  2  Tr\left (K G W \right )
+ Tr\left ( K G W W^\top G^\top \right ) \right ] \\
&+ \alpha \frac{1}{2} \sum_{i,j=1}^n \left ( \sum_{k=1}^l \mu_k K^k_{ij} \right ) \left \| \bfw_i - \bfw_j \right \|_2^2 \\
s.t.
~& H \geq 0, G \geq 0,
\end{aligned}
\end{equation}
which can be solved by Lagrange multiplier method.

\item \textbf{Updating $\{\mu_k\}_{k=1,\cdots,l}$  while fixing $H$ and $G$}:  When $H$ and $G$ are fixed, the problem in (\ref{equ:MultiKGNMF}) is reduced to

\begin{equation}
\begin{aligned}
\min_{\{\mu_k\}_{k=1,\cdots,l}}
~ & \sum_{k=1}^l \mu_k \left [ Tr\left ( K \right ) -  2  Tr\left (K G W \right )
+ Tr\left ( K G W W^\top G^\top \right ) + \alpha \frac{1}{2} \sum_{i,j=1}^n  K^k_{ij} \left \| \bfw_i - \bfw_j \right \|_2^2 \right ]\\
& + \beta \sum_{k=1}^l \mu_k^2\\
s.t.
~& \sum_{k=1}^l \mu_k= 1,  \mu_k \geq 0, k=1,\cdots,l,
\end{aligned}
\end{equation}
and it can also be solved as a QP problem.

\end{itemize}

\section{Summary}
\label{sec:summary}

In this chapter, we investigate a fundamental problem in manifold regularized NMF methods --- how to construct an optimal graph to present the manifold. We introduce three different methods to this problem, which consider multiple graph learning, feature selection, and multiple kernel learning methods, and integrate them to the problem of NMF. The common features of these methods are of two folds:

\begin{enumerate}
\item The parameters of multiple graph learning, feature selection, and multiple kernel  are all learned according to the criterion of NMF, and no supervision information is needed. This is critical for many unsupervised learning methods.
\item The parameters of NMF, multiple graph learning, feature selection, and multiple kernel learning are coupled in a single objective function, but we employed alternate optimization methods to update them. Interestingly, we found that the parameters of multiple graph learning, feature selection, and multiple kernel learning can all be optimized by solving QP problems.
\end{enumerate}

Although the methods proposed in this chapter is to regularize the learning of NMF, it provides some insights to manifold learning. Traditional manifold learning methods all use a simple method to construct a nearest neighbor graph and use it to regularize different forms of outputs. However, the construction of the graph itself has not attracted much attention. In this chapter, we discussed how to construct an optimal graph from different affinity measures, noisy features, and different kernels. In the future, we will extend the work in this chapter to different learning problems, such as sparse coding \cite{wang2014semi,al2014supervised,wang2013feature,wang2013discriminative}, learning to rank \cite{wang2014sparse}, classification \cite{wang2014large}, etc. Moreover, we will also investigate adapting the proposed manifold learning methods to big data using distributed systems \cite{6628011,6584814,wang2013supporting,wang2012scimate,wang2012scimate,wang2010mfti,lim2010ultra,lei2010decade,zhang2012reliable}, and applying these methods to applications of sensing \cite{LiuTem2012a}, computer vision \cite{6619001,6607541,shen2013virtual,zhou2013adaptive,zhou2013adaptive}, pattern recognition \cite{wang2015maximum,luo2009parameter,brodsky2008corejava}, signal processing \cite{wang2012passivity,wang2010peds,lei2010vector}, networking \cite{Xu2012TAAS,cross13,yang2014location,web14}, and hardware fault detection \cite{wang2010hardware,yu2013addressing,yu2011exploiting,li2013reliable,ampadu2013breaking,zhang2014two,zhang2013variation,zhang2012fine},

\end{document}